\documentclass[10pt,compsoc,journal,cspaper]{IEEEtran}
\usepackage{graphicx}    
\usepackage{calrsfs}
\usepackage{subfigure}
\usepackage{booktabs}
\usepackage{colortbl}
\usepackage{tabularx}
\usepackage{setspace}
\usepackage{color}
\usepackage{algorithm2e}
\usepackage{amsmath}
\usepackage{epsfig}
\usepackage{bm}
\makeatletter
\newcommand{\nosemic}{\renewcommand{\@endalgocfline}{\relax}}
\newcommand{\dosemic}{\renewcommand{\@endalgocfline}{\algocf@endline}}
\makeatother
\DeclareMathOperator*{\argmin}{argmin} 
\begin{document}

\author{Masud~Moshtaghi,
James~C.~Bezdek,
Sarah~M.~Erfani, 
Christopher~Leckie,
James~Bailey
\IEEEcompsocitemizethanks{\IEEEcompsocthanksitem M. Moshtaghi, J.C. bezdek, S. Erfani, C. Leckie and J.Bailey were with the School of Computing and Information Systems, University of Melbourne, Melbourne, Australia, 3010. M. Moshtaghi is now at Amazon.com \protect\\
E-mail: masud.moshtaghi@unimelb.edu.au}%
\thanks{}}


\title{Online Cluster Validity Indices for \\ Streaming Data}

\IEEEtitleabstractindextext{
\begin{abstract}
Cluster analysis is used to explore structure in unlabeled data sets in a wide range of applications. An important part of cluster analysis is validating the quality of computationally obtained clusters. A large number of different internal indices have been developed for validation in the offline setting. However, this concept has not been extended to the online setting. A key challenge is to find an efficient incremental formulation of an index that can capture both cohesion and separation of the clusters over potentially infinite data streams. In this paper, we develop two online versions (with and without forgetting factors) of the Xie-Beni and Davies-Bouldin internal validity indices, and analyze their characteristics, using two streaming clustering algorithms (sk-means and online ellipsoidal clustering), and illustrate their use in monitoring evolving clusters in streaming data. We also show that incremental cluster validity indices are capable of sending a distress signal to online monitors when evolving clusters go awry. Our numerical examples indicate that the incremental Xie-Beni index with forgetting factor is superior to the other three indices tested.
\end{abstract}}
\maketitle

\section{Introduction}
The intrinsic nature of streaming data requires algorithms that are capable of fast data analysis to extract knowledge. Online clustering algorithms provide a way to extract patterns from continuous data streams. Therefore, online clustering has gained popularity in applications involving massive streams of data, such as router packet analysis and environmental sensing~\cite{Guha2003,Silva2013}. In these applications, the velocity and volume of data is too high for the processing unit to access each data sample more than once. A category of fast online clustering algorithms, also referred to as sequential clustering, process data quickly and efficiently by receiving samples one at a time and updating cluster statistics (such as prototypes) with each sample~\cite{Ackerman2014, Angelov2010, OEC2016}.

An important aspect of any clustering algorithm is assessment of the quality of the resulting clusters, i.e., how well does any partition match the input data?
\textit{Cluster validity indices} (CVIs) comprise computational models and algorithms whose job is to identify the ``best" member among different partitions of the input data. Most CVIs are max-optimal or min-optimal, meaning that the partition preferred by the index is indicated by the maximum (minimum) value of the index on the partitions being evaluated. To date, CVIs have been used exclusively in a static (or batch) setting, being applied to sets of partitions generated by different parameter settings of the clustering algorithm applied to a collected set of data. In online clustering, the question of how well a clustering algorithm matches the data becomes even more important as we need to know how the clustering algorithm is performing at any point in time and how it reacts to the changes in the partitions over time. The key assumption in the online context is that data are \textit{processed once}, and \textit{historical data will not be available} for a retrospective analysis. 

While batch clustering algorithms are supported by a wide range of cluster validation methods~\cite{Arbelaitz2013}, no effort has been made to extend cluster validation methods to online clustering algorithms. 
This study concerns itself with the use of internal \textit{incremental cluster validity indices} (iCVIs) (we drop the word ``internal" in the rest of the paper), that are computed online (on the fly) and used to control/interpret clustering for streaming data.

Most of existing batch indices assess two basic characteristics of a set of clusters: compactness (or density) and separation of the clusters~\cite{Arbelaitz2013}. Compactness is usually calculated based on the observations while separation is often measured by the distance between cluster prototypes. In the online setting, where each observation can be accessed only once, an incremental/recursive calculation of compactness is necessary. In this paper, we propose two incremental methods to estimate within cluster dispersion: (1) an exact incremental update of a batch formula; and (2) an online formula incorporating an exponential forgetting factor. Using these two methods we then derive online versions of two well-known CVIs namely, the \textit{Xie-Beni} (XB)~\cite{Xie91} and the \textit{Davies-Bouldin} (DB)~\cite{Davies1979} indices. These indices can be applied to both hard and soft partitions.

This paper offers four main contributions: (1) we propose a new concept of incremental validation of online clustering algorithms which provides new insights into these algorithms; (2) we propose two incremental versions of within cluster dispersion, viz., with and without forgetting, a facility that enables the iCVI to gracefully forget earlier inputs; (3) we propose incremental versions of two well-known batch CVIs allowing exact calculation of the two indices with fast sequential processing of data; and (4) we analyze and discuss the properties of the proposed iCVIs within the context of two online clustering algorithms. Our results demonstrate useful insights produced by the iCVIs about the partitions produced by online clustering algorithms. Moreover, the proposed iCVIs can indicate learning difficulties experienced by the clustering algorithm and can signal the appearance of new clusters in the evolution of the data stream. 

The next section summarizes related work. In Section~\ref{sec:problemstatement}, we present definitions and notation needed in this paper.  Section~\ref{sec:ClustAlgs} contains background information on two important online clustering algorithms. In Section~\ref{sec:iCVI}, we derive two versions of the iXB, iDB indices and analyze their characteristics. Section~\ref{sec:evaluation1} contains numerical examples used to evaluate the proposed models. A summary and conclusions are given in Section~\ref{sec:conclusion}.

\section{Background and Related Work}
\label{sec:relatedwork}
In this section, we briefly describe related work in cluster validation and online clustering algorithms. It is especially important to record our definition of an online clustering algorithm as our goal is to validate these algorithms using cluster validity indices.

According to Guha et al.~\cite{Guha2003}, clustering in data streams can be divided into two main strategies: (1) buffering a window of streamed inputs,  finding clusters in the window using a batch algorithm such as the classic k-means algorithm; and then merging clusters in adjacent windows to obtain a final clustering. This strategy is espoused, for example, in~\cite{Aggarwal2007,Ailon2009,PAKDD2014}; (2) using incremental learning techniques to find clusters in the evolving data stream. Examples of this strategy include~\cite{Ackerman2014,OEC2016,Angelov2008,Cao2006,Kranen2011,Mozafari2014}. We refer to this second approach as online or incremental clustering. Algorithms for online clustering can themselves be divided into two categories. The first category is general clustering algorithms for any sequence of data (we call this \textit{type 2a} data). These algorithms do not assume any ordering in the data stream and require the number of clusters to be specified in advance, for example sequential k-means, or more briefly, sk-means, and sequential agglomerative clustering~\cite{Ackerman2014}. A second category of online clustering algorithms assume a natural ordering in the data (time-series) and operate on the assumption that close observations in time will be closely related. These algorithms use this assumption to dynamically create clusters in evolving data streams (we call this \textit{type 2b} data). In this paper, we propose cluster validation methods for online clustering algorithms of types 2a and 2b.

Cluster validity indices (CVIs) can be grouped into two categories: \textit{internal} and \textit{external} indices. Internal indices use only the information available from the
algorithmic outputs and the observed unlabeled data. In contrast, external CVIs use additional external information about substructure in the data, usually in the form of a reference partition (a ground truth partition), so the data processed are labeled. External CVIs are used to compare partitions obtained by a clustering algorithm to ground truth labels. Another use of external CVIs is to correlate external and internal assessments of labeled data~\cite{Arbelaitz2013}. In this application the external CVI becomes a tool for selection of the ``best" internal CVI when unlabeled data are to be clustered. The use of external CVIs to choose a ``good" internal CVI is comprehensively discussed in~\cite{Arbelaitz2013}. Our focus in this paper is on internal CVIs. Milligan and Cooper's 1985  paper is widely regarded as a landmark study for the comparison of internal CVIs~\cite{Milligan1985}. 

Internal cluster validity indices fall under the broad umbrella of goodness-of-fit measures such as likelihood ratio tests and Root Mean Squared Error~\cite{Schermelleh-Engel2003}. The majority of goodness-of-fit measures target parametric models while CVIs provide a non-parametric mechanism to evaluate clustering outputs. Another differentiating point between internal CVIs and goodness-of-fit measures is the meaning of the fitness term. Most internal CVI models have components that attempt to capture cohesion and separation, while goodness of fit indices usually assess the fit of a model to the data that generates it. 

There are two categories of internal CVIs based on the way that they measure cohesion and separation. The CVIs in the first category use only the partitions generated by the clustering to determine the quality of the partition. Measures of this type include the partition coefficient and partition entropy~\cite{Bezdek1981}. Indices such as these often appear in the context of fuzzy cluster validity. However, most CVIs fall into the second category, that is, they use both the data and the partition information.

We develop incremental CVIs by deriving an incremental formula for the data-dependent part of two well-known indices, i.e., XB and DB. After determining the incremental update formula for cluster cohesion, we derive one step update formulae for these two indices. We then investigate their application to cluster analysis in data streams. While the results of any online clustering algorithm can be analyzed using the proposed iCVIs, we focus on two clustering algorithms - a crisp clustering algorithm from the general data stream clustering category (type 2a), viz., sk-means; and the \textit{online elliptical clustering} (OEC) clustering algorithm from the time-series clustering category (type 2b), which is a soft/fuzzy clustering algorithm.


\section{Problem Statement and Definitions}
\label{sec:problemstatement}

Traditional batch clustering algorithms aim to find crisp or fuzzy/probabilistic \textit{k}-partitions of a collection of $n$ samples of \textit{static} data, viz., $X=\left\{\bm{x}_1,\bm{x}_2,\ldots,\bm{x}_{n},\ldots,\bm{x}_n\right\}\subset\Re^p$. All vectors in this article are column vectors. Crisp and fuzzy partitions of $X$ are conveniently represented by matrices in the following sets:
\begin{equation}
\begin{split}
M_{\textit{fkn}}= & \left\{\right.U\in\Re^{kn}: \textit{for } 1\leq i \leq k,1\leq j \leq n:\\ & 0\leq u_{ij} \leq 1:
\sum_{i=1}^{k}u_{ij}=1\;\forall j;\;\sum_{j=1}^{n}u_{ij}>0\;\forall i\left.\right\};
\end{split}
\label{eq:FuzzyPartition1}
\end{equation}
\begin{equation}
M_{\textit{hkn}}= \left\{U\in M_{\textit{fkn}}: u_{ij}\in\left\{0,1\right\}\forall i,j\right\}.
\label{eq:FuzzyPartition2}
\end{equation}

Now suppose that $n$ inputs have arrived sequentially in the streaming data, and we have found, from these $n$ inputs $U_{n}\in M_{\textit{fkn}}$, a set of soft/fuzzy clusters of $X$, together with a set $V_{n}=\left\{\bm{v}_{1n},\ldots,\bm{v}_{kn}\right\}\subset\Re^{kp}$ of cluster centers. We can use $\left(U_{n},V_{n}\right)$ to calculate various CVIs.

When input $\bm{x}_{n+1}$ arrives, it is used by an online clustering algorithm to find the membership, $u_{i,n+1}$, of the new point in the $i^{\textit{th}}$ cluster, $1\leq i\leq k$. Let the vector $\bm{u}_{n+1}=\left\{u_{i,n+1}| i=1,\ldots,k\right\}$ be the label vector of $\bm{x}_{n+1}$ in the set of ($k$) clusters. The clustering algorithm will also use $\bm{x}_{n+1}$ to update $V_{n}\rightarrow V_{n+1}$. 

Given the new input $\bm{x}_{n+1}$, its cluster assignment $\bm{u}_{n+1}$ and the updated cluster centers $V_{n+1}$, we will derive one-step update formula for a particular CVI.

\begin{figure}[ht]
	\centering
			\includegraphics[width=8.3cm]{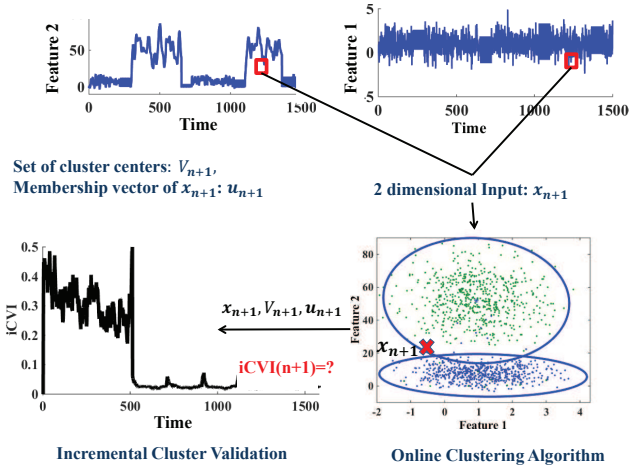}
		\caption{One-step update of incremental CVI in an online setting.}
	\label{fig:problemprocess}
\end{figure}

The calculation of $\bm{u}_{n+1}$ and updates to $V$ are done using specific incremental clustering algorithms such as sk-means~\cite{Ackerman2014} or OEC~\cite{OEC2016}. The question posed here is how the value of the chosen CVI changes incrementally with this update. Fig.~\ref{fig:problemprocess} illustrates the overall process. The two time series at the top of the figure form the input to the online clustering algorithm on the bottom right. When $\bm{x}_{n+1}$ becomes available, the online clustering algorithm finds the membership values $\bm{u}_{n+1}$, and updates the cluster centers to produce $V_{n+1}$. Then, $\bm{u}_{n+1}, V_{n+1}, \bm{x}_{n+1}$ are passed to the incremental cluster validation process. The objective in this paper is to answer the question ``iCVI(n+1)=?" posed in the bottom left panel of Fig.~\ref{fig:problemprocess}.

To answer this question, we first need to describe how $\bm{u}_{n+1}$ and $V_{n+1}$ are calculated incrementally by online clustering algorithms. Section~\ref{sec:ClustAlgs}, briefly describes two algorithms to solve this problem.   



\section{Incremental Clustering Algorithms}
\label{sec:ClustAlgs}
The calculation of iCVIs depends on the information provided by a particular clustering algorithm. Therefore, we start by providing a brief overview of two different types of incremental clustering algorithms.

\subsection{Sequential k-means}
The sk-means algorithm shown below as Algorithm~\ref{alg:SeqKmeans} and its variants have been studied in various forms in the literature of  \textit{self organizing maps} (SOMs)~\cite{Ackerman2014,Pal1996}. Algorithm~\ref{alg:SeqKmeans} records the basic sequential k-means method studied by Macqueen~\cite{Macqueen67}.
\begin{algorithm}[ht]
\singlespacing
\SetKwInOut{Input}{Input}
\SetKwInOut{Note}{Note}
\KwData{$X$ - set of data points}
\Input{$k$ - number of clusters}
\Note{$\left\|.\right\|$ is the Euclidean norm}
Initialize $V_k$ with the first $k$ data points $V_k=\left\{\bm{x}_1,\bm{x}_2,\ldots,\bm{x}_k\right\}$\;
$U_k$ is the $k\times k$ identity matrix\;
Initialize the counter for each cluster $\left\{n_1,n_2,\ldots,n_k\right\}$ with 1\;
\ForEach{$\bm{x}_{n}$ in the stream}{
$m=\argmin_{m\in\left\{1\ldots k\right\}}\left\|\bm{x}_{n}-\bm{v}_m\right\|$\;
$n_m=n_m+1$\;
$\bm{v}_m=\bm{v}_m+\left(\bm{x}_{n}-\bm{v}_m\right)/n_m$\;
$\bm{u}_{n}=\left(0,0,\ldots,1,\ldots,0\right)^T\in\left\{0,1\right\}^k$, where the 1 appears in the $m-th$ place, is appended to the current $U$ as its $n-th$ column\;
}
\caption{The basic sequential k-means algorithm (Macqueen~\cite{Macqueen67}).}
\label{alg:SeqKmeans}
\end{algorithm}

Macqueen's algorithm requires pre-specification of $k$, the number of clusters it builds. After initializing the $k$ clusters by designating the first $k$ points in $X$ as the initial cluster centers, when the next input arrives, sk-means computes the distance from the next input to the $k$ cluster centers, and assigns the input point to the cluster of the nearest prototype. Then the winning prototype is updated as shown in the third line of the \textit{for} part of Algorithm~\ref{alg:SeqKmeans}. So at step $n+1$ of the algorithm the $k$ vector $\bm{u}_{n+1}$ has only one element with the value of 1 (the winner of a nearest prototype competition), and the remaining $k-1$ values are 0.

\subsection{Online Ellipsoidal Clustering}
\label{sec:OEC}
This section briefly reviews the \textit{online ellipsoidal clustering} (OEC) algorithm developed in~\cite{OEC2016}. This clustering algorithm is specific to time-series and does \textit{not} require the number of clusters to be pre-specified. This algorithm has ellipsoidal cluster prototypes defined by $k$ sets of  means and covariance matrices $\left\{(\bm{m}_{i,n},S^{-1}_{i,n})|i=1,\ldots,k\right\}$ at any step $n$. Two different radii are considered for each ellipsoid, one called the \textit{effective boundary}, i.e., the smaller radius, and the other one is called the \textit{outlier boundary}. The outlier boundary prevents the cluster prototype from being affected by large outliers. These boundary thresholds are selected from the inverse of the chi-squared distribution, $(\chi^{2})^{-1}_{p}(\gamma)$, where $\gamma$ is the probability that a cluster member falls inside the ellipsoid.
 In the OEC clustering algorithm, cluster center $\bm{v}_{i,n+1}$ is formed from the sample mean $\bm{m}_{i,n+1}$ of each cluster, and $\bm{u}_{n+1}$ is formed by the values obtained from the soft assignments of the input point to the clusters, $\bm{u}_{i,n+1},\;1\leq i\leq k$, defined as

\begin{equation}
\bm{u}_{i,n+1} = \left[\sum_{j=1}^{k}\left(F_{i,n+1}/F_{j,n+1}\right)^{2/(m-1)}\right]^{-1},\;\;\; m\in(1,\infty),
\label{eq:FCMWeight}
\end{equation}

\noindent where $F_{i,n+1}=(\bm{x}_{n+1}-\bm{m}_{i,n})^{T}S^{-1}_{i,n}(\bm{x}_{n+1}-\bm{m}_{i,n})$. Readers may recognize this as the membership update formula required by fuzzy k-means using the sample-based Mahalanobis norm for the inner product induced distance~\cite{Bezdek2017}. The OEC algorithm uses two other parameters that we need to set during the experiments. 

\noindent\textit{Stabilization period ($n_s$) - } An ellipsoidal prototype in $\Re^{p}$ is created by $p+1$ consecutive distinct points, but to obtain a reliable estimate of $(\bm{m}_{i,n},S^{-1}_{i,n})$, more data from the input stream is required. The integer $n_s$ stabilizes these incremental estimates by temporarily disabling the OEC guard zone and new cluster detection tests until the current cluster contains $n_s$ points. 

\noindent\textit{Forgetting factor ($\lambda_{\textit{OEC}}$)- } OEC has a special forgetful prototype. When this prototype does not overlap with any of the existing clusters in the system, a new cluster is formed. This prototype use an exponential forgetting factor ($\lambda_{\textit{OEC}}$) for this purpose.

Now that we have described how $\bm{v}_{n+1}$ and $\bm{u}_{n+1}$ can be updated using two different clustering algorithms, we can describe how iCVIs use this information to assess the evolving clusters incrementally. In Section~\ref{sec:iCVI}, we introduce an incremental calculation of a common cohesion measure in CVIs, then we discuss how two particular CVIs which use this cohesion measure can be updated using its previous value at time $n$ with the additional information provided by $\bm{u}_{n+1},V_{n+1}$ and $\bm{x}_{n+1}$.

\section{The Incremental CVIs}
\label{sec:iCVI}
Normally there is no external ground truth information available with streaming inputs, hence we explore the usage of internal CVIs to assess the evolving performance of online clustering algorithms that generate both memberships and cluster prototypes. We consider two well-known indices in this group that work for both hard and soft partitions. 

The first index of this type is the \textit{Xie-Beni} (XB) index~\cite{Xie91}. A general formulation of this index for a batch collection of $n$ inputs is~\cite{Pal1995}, where $\mathcal{A}$ is a positive-definite weight matrix which induces the inner product norm $\left\|x\right\|^{2}_\mathcal{A}=x^T\mathcal{A}x$:
\begin{equation}
\begin{split}
& \textit{XB}_{m\mathcal{A}}\left(U,V;X\right)=\frac{J_{m\mathcal{A}}\left(U,V;X\right)}{n\left(\min_{i\neq j}\left\{\left\|\bm{v}_i-\bm{v}_j\right\|_{\mathcal{A}}^{2}\right\}\right)}\\
& \text{, where }m\in[1,\infty)\text{, and }\\
& J_{m\mathcal{A}}\left(U,V;X\right)= \sum_{j=1}^{n}\sum_{i=1}^{k}(u_{ij})^m\left\|\bm{x}_{n}-\bm{v}_i\right\|_{\mathcal{A}}^{2}.
\end{split}
\label{eq:XB}
\end{equation}

\noindent The parameter $m$ is called the fuzzifier of the model. For simplicity, we consider only $\mathcal{A}=I_p$ (the Euclidean norm)and $m=2$, and drop subscripts, writing $J_{2I_{p}}$ as $J$ in the sequel.

The second index is a relative of the DB index~\cite{Davies1979} introduced by Araki~\emph{et al.}~\cite{Araki1993},

\begin{equation}
\begin{split}
& \textit{DB}\left(U,V;X\right)=\frac{1}{k}\sum_{i=1}^{k}{\max_{j,j\neq i}{\frac{L_{i}+L_{j}}{\left\|\bm{v}_i-\bm{v}_j\right\|^2}}}\\
& L_i = \frac{\sum_{j=1}^{n}{u_{ij}^2\left\|\bm{x}_{j}-\bm{v}_{i}\right\|^2}}{\sum_{j=1}^{n}{u_{ij}^2}}.
\end{split}
\label{eq:DBI}
\end{equation}

\noindent This is not a true generalization of the the DB index because square roots are missing for this choice of $p=q=2$ in ~\cite{Davies1979}.  but the values of (\ref{eq:DBI}) are closely related to the true DB index in the crisp case. The common part of these two (and many other) indices is the way that they capture the within-cluster dispersion. We define the \textit{fuzzy within cluster dispersion}, for $U\in M_{fkn}$, as,

\begin{equation}
\begin{split}
C_{i,n}=\sum_{j=1}^{n}{(u_{ij})^2\left\|\bm{x}_{j}-\bm{v}_{i,n}\right\|^2}.
\end{split}
\label{eq:icd}
\end{equation}

This is the only part of either index that depends directly on the input data. Therefore, we first develop an incremental calculation of this part of the index, and then use it to propose formulae for incremental calculation of these two indices. 


\subsection{Incremental Cluster Dispersion Measure}
In this section, we derive a formula for (\ref{eq:icd}) at time step $n+1$ based on its value at time step $n$. We assume that at time $n+1$, all of the previous input values $\bm{x}_1, ..., \bm{x}_n$ have been discarded, so the only data point we have to work with is $\bm{x}_{n+1}$.The goal is to write the update formula in terms of the value in the previous step and a change to this value on seeing the $n+1^{\textit{st}}$ input, i.e., $C_{i,n+1}=C_{i,n} + \Delta C_{i,n}$. We begin with

\begin{equation}
\begin{split}
C_{i,n+1}=\sum_{j=1}^{n+1}{(u_{ij})^2\left\|\bm{x}_{j}-\bm{v}_{i,n+1}\right\|^2}
\end{split}
\label{eq:deltaC}
\end{equation} 

First we isolate the effect of the last point in the summation to obtain

\begin{equation}
\begin{split}
C_{i,n+1} & = \sum_{j=1}^{n}{(u_{ij})^2\left\|\bm{x}_{j}-\bm{v}_{i,n+1}\right\|^2}+ \\
& \underbrace{(u_{i,n+1})^2\left\|\bm{x}_{n+1}-\bm{v}_{i,n+1}\right\|^2}_{A_{i,n+1}}.
\end{split}
\label{eq:CApprox}
\end{equation}

At this point if we could assume that $\left\|\bm{v}_{i,n+1}-\bm{v}_{i,n}\right\|\approx 0$, we would have $C_{i,n+1}=C_{i,n}+A_{i,n+1}$. Here, we are looking for an exact calculation of $C_{i,n+1}$ so we need to compute the effect of the change in the cluster centers. To isolate the change of the centers, we add and subtract $\bm{v}_{i,n}$ inside the Euclidean norm in the first term in (\ref{eq:CApprox}), and rewrite the norm in terms of the (Euclidean) inner product,

\begin{equation}
\begin{split}
& C_{i,n+1}= \sum_{j=1}^{n}{(u_{ij})^2<\bm{x}_{j}-\bm{v}_{i,n}+\bm{v}_{i,n}-\bm{v}_{i,n+1},\bm{x}_{j}-\bm{v}_{i,n}} \\
& + \bm{v}_{i,n}-\bm{v}_{i,n+1}> + A_{i,n+1}.
\end{split}
\label{eq:CNorms1}
\end{equation}

\noindent After few steps and some simplification we obtain 
\begin{equation}
\begin{split}
& C_{i,n+1}= C_{i,n} + \overbrace{A_{i,n+1}+ M_{i,n}B_{i,n+1}+2Q_{i,n+1}}^{\Delta C_{i,n}}
\end{split}
\label{eq:CNorm1}
\end{equation} 
\noindent where
\begin{equation}
\begin{split}
& Q_{i,n+1}=\sum_{j=1}^{n}{(u_{ij})^2\left\langle\bm{x}_{j}-\bm{v}_{i,n},\bm{v}_{i,n}-\bm{v}_{i,n+1}\right\rangle}
\end{split}
\label{eq:Q}
\end{equation}
\begin{equation}
\begin{split}
& B_{i,n+1} = \left\|\bm{v}_{i,n}-\bm{v}_{i,n+1}\right\|^2
\end{split}
\label{eq:B}
\end{equation}

\begin{equation}
\begin{split}
& M_{i,n+1}=M_{i,n}+u^{2}_{i,n+1},\;M_{i,1}=u^{2}_{i,1}
\end{split}
\label{eq:M}
\end{equation}

\noindent The term $Q_{i,n+1}$ in equation (\ref{eq:Q}) depends on the previous (discarded) values of $\bm{x}_{j}$ and $u_{ij}$ where $j=1,\ldots,n$, so we cannot yet make the incremental calculation of $\Delta C_{i,n}$. Since the second part of the dot product in (\ref{eq:Q}), i.e., $\left(\bm{v}_{i,n}-\bm{v}_{i,n+1}\right)$, does not depend on $j$, we can write $Q_{i,n+1}$ as

\begin{equation}
\begin{split}
Q_{i,n+1}=\left[\bm{v}_{i,n}-\bm{v}_{i,n+1}\right]^{T}\overbrace{\left[\sum_{j=1}^{n}{(u_{ij})^2\left(\bm{x}_{j}-\bm{v}_{i,n}\right)}\right]}^{G_{i,n}}
\end{split}
\label{eq:Q1}
\end{equation}

Using the same trick of adding and subtracting $\bm{v}_{i,n}$ in (\ref{eq:Q1}) we can write an incremental update formula for $G_{i,n+1}$,

\begin{equation}
\begin{split}
& G_{i,n+1}= G_{i,n} + \Delta G_{i,n}, G_{i,1}=\overrightarrow{0} \\
& \Delta G_{i,n} = M_{i,n}(\bm{v}_{i,n}-\bm{v}_{i,n+1})+(u_{i,n+1})^2\left(\bm{x}_{n+1}-\bm{v}_{i,n+1}\right).
\end{split}
\label{eq:IncG}
\end{equation}

We now have all the components needed to calculate $\Delta C_{i,n}$. The two terms $A_{i,n+1}$ and $B_{i,n+1}$ are calculated directly and $Q_{i,n+1}$ and $M_{i,n+1}$ are incrementally calculated. The Algorithm~\ref{alg:Compactness} depicts a function that incrementally calculates the compactness.

\begin{algorithm}[ht]
\singlespacing
\SetKwInOut{Input}{Input}
\SetKwInOut{Output}{Output}
\SetKwComment{Comment}{/* }{}
\KwData{$\bm{v}_{i,n},\bm{v}_{i,n+1},u_{i,n+1}$, $\bm{x}_{n+1}$}
\Input{$G_{i,n}, M_{i,n}, C_{i,n}$}
\Output{$G_{i,n+1}, M_{i,n+1}, C_{i,n+1}$}
\Comment{note $i=1,\ldots,k$}
 \ForEach{$i \in \left\{1,\ldots,k\right\}$}{   
		$Q_{i,n+1} = \left[\bm{v}_{i,n}-\bm{v}_{i,n+1}\right]^{T}G_{i,n}$\;
    $B_{i,n+1} = \left\|\bm{v}_{i,n}-\bm{v}_{i,n+1}\right\|^2$\;
	  $A_{i,n+1} = (u_{i,n+1})^2\left\|\bm{x}_{n+1}-\bm{v}_{i,n+1}\right\|^2$\;
    $C_{i,n+1} =  C_{i,n}+A_{i,n+1}+M_{i,n}B_{i,n+1}+2Q_{i,n+1}$\;
    \nosemic $G_{i,n+1}= G_{i,n}+M_{i,n}(\bm{v}_{i,n}-\bm{v}_{i,n+1})+$\;
		\Indp\Indp\Indp\dosemic $(u_{i,n+1})^2\left(\bm{x}_{n+1}-\bm{v}_{i,n+1}\right)$\;
    \Indm\Indm\Indm$M_{i,n+1} = M_{i,n}+u^2_{i,n+1}$\;
}
\caption{The incremental compactness function for data point $\bm{x}_{n+1}$.}
\label{alg:Compactness}
\end{algorithm}
%

As $n\rightarrow\infty$ the effect of each new sample on the total value of $C_{i,n}$ is expected to become small. In data streaming applications, which have, in theory, an infinite data stream, we are interested in how well a clustering algorithm keeps up with the evolution of cluster structure of points in the stream. Our objective is to use iCVIs to capture the quality of the partitions over a window of the most recent observations at any point in time.

Exponential fading memory is a common approach in online learning methods. In this approach a forgetting factor $0<\lambda<1$ is included in the incremental estimations so that the data sample from $f$ steps before the current sample is weighted by $\lambda^f$. In this way, older samples become less and less relevant to the current estimation. The batch representation of $C_{i,n}$ with a forgetting is shown in (\ref{eq:BIncidff}).

\begin{equation}
C_{\lambda i,n}=\sum_{j=1}^{n}{\lambda^{n-j}u_{ij}^2\left\|\bm{x}_{j}-\bm{v}_{i,n}\right\|^2} 
\label{eq:BIncidff}
\end{equation}

An argument similar to the one used to derive equation (\ref{eq:CNorm1}) leads to an incremental update formula for $C_{\lambda i,n+1}$.

\begin{equation}
\begin{split}
& Q_{\lambda i,n+1}=\left(\bm{v}_{i,n}-\bm{v}_{i,n+1}\right)G_{\lambda i,n}
\end{split}
\label{eq:IncQFF}
\end{equation}

\begin{equation}
\begin{split}
B_{i,n+1} = \left\|\bm{v}_{i,n}-\bm{v}_{i,n+1}\right\|^2
\end{split}
\label{eq:IncBFF}
\end{equation}

\begin{equation}
\begin{split}
A_{i,n+1} = u^2_{i,n+1}\left\|\bm{x}_{n+1}-\bm{v}_{i,n+1}\right\|^2
\end{split}
\label{eq:IncAFF}
\end{equation}

\begin{equation}
\begin{split}
& C_{\lambda i,n+1} = \lambda C_{\lambda i,n}+ \Delta C_{\lambda i,n} \\
& \Delta C_{\lambda i,n} = 2\lambda Q_{\lambda i,n+1}+\lambda M_{\lambda i,n}B_{i,n+1}+A_{i,n+1}
\end{split}
\label{eq:IncicdFF}
\end{equation}

\begin{equation}
\begin{split}
& G_{\lambda i,n+1}= \lambda G_{\lambda i,n}+ \Delta G_{\lambda i,n} \\
& \Delta G_{\lambda i,n} = \lambda M_{\lambda i,n}(\bm{v}_{i,n}-\bm{v}_{i,n+1})+(u_{i,n+1})^2\left(\bm{x}_{n+1}-\bm{v}_{i,n+1}\right).
\end{split}
\label{eq:IncGFF}
\end{equation}

\begin{equation}
\begin{split}
M_{\lambda i,n+1}= \lambda M_{i,n}+ u_{i,n}^2,M_{i,1\lambda}=u_{i,1}^2
\end{split}
\label{eq:IncMFF}
\end{equation}

In the next section, we use these formulas to derive two incremental versions of the XB and DB indices at equations (\ref{eq:XB}) and (\ref{eq:DBI}). 

\subsection{Incremental Xie-Beni Index}
Let XB($n+1$) denote the value of the XB index we seek when $\bm{x}_{n+1}$ arrives after $n1$ inputs have been processed. Let $J_{n+1}$ denote the value of $J$ at step $n+1$.
To compute XB incrementally, i.e., to compute XB($n+1$), we need an incremental update for $J_{n+1}$ and for the denominator of (\ref{eq:XB}). The numerator of XB is updated using the value of $C_{i,n+1}$ from Algorithm~\ref{alg:Compactness}, and the denominator is calculated with the updated centers $V_{n+1}$.  Equation (\ref{eq:JmUpdate}) shows the one step update of $J_{n+1}$ at step $n+1$.

\begin{equation}
J_{n+1}=\sum_{i=1}^{k}{C_{i,n+1}}.
\label{eq:JmUpdate}
\end{equation}
Let 
\begin{equation}
h_{n+1}=\min_{i\neq j,\bm{v}_{i},\bm{v}_j\in V_{n+1}}\left\{\left\|\bm{v}_i-\bm{v}_j\right\|^2\right\}. 
\label{eq:CN}
\end{equation}
Then value of the incremental XB index, is 
\begin{equation}
\begin{split}
\textit{XB}(n+1)= \frac{J_{n+1}}{(n+1)h_{n+1}}.
\end{split}
\label{eq:iXB}
\end{equation}

For batch clustering, the case $k=1$ is usually not considered. However, in the streaming environment and in algorithms like OEC, the number of clusters dynamically changes and starts from $k=1$. With one cluster, $h_{n+1}$ is undefined in (\ref{eq:CN}). When $k=1$, we replace (\ref{eq:CN}) with $h_{n+1}=\max{\left\{h_{n},\left\|\bm{v}_1-\bm{x}_{n+1}\right\|^2\right\}}$. 

The one step update of iXB with forgetting is obtained by replacing $C_{i,n+1}$ with $C_{\lambda i,n+1}$ in (\ref{eq:JmUpdate}) to obtain in $J_{\lambda,n+1}$,  
\begin{equation}
\begin{split}
\textit{XB}_{\lambda}(n+1)= \frac{(1-\lambda)J_{\lambda,n+1}}{h_{n+1}}.
\end{split}
\label{eq:iXBFF}
\end{equation}

Please note that XB($n+1$) and XB$_{\lambda}$($n+1$) are the values of the incremental XB indices without and with the forgetting factor after $\bm{x}_{n+1}$ is processed, while iXB and iXB$_{\lambda}$ are the names of the incremental XB models.
\subsection{Incremental DB Index}
Let DB($n$) denote the value of the Davies-Bouldin index after $n$ inputs. We want to compute incrementally updated values DB($n+1$) and DB$_\lambda$($n+1$) when input $\bm{x}_{n+1}$ arrives. We need to normalize $C_{i,n+1}$ and $C_{\lambda i,n+1}$ with the number of data points in the $i^{\textit{th}}$ cluster. The index DB($n+1$) at time step $n+1$ can be written as

\begin{equation}
\begin{split}
\textit{DB}(n+1)=\frac{1}{k}\sum_{i=1}^{k}{\max_{j,j\neq i}{\frac{L_{i,n+1}+L_{j,n+1}}{\left\|\bm{v}_{i,n+1}-\bm{v}_{j,n+1}\right\|^2}}},
\end{split}
\label{eq:incDBI}
\end{equation}

\noindent where 

\begin{equation}
\begin{split}
L_{i,n+1} = \frac{C_{i,n+1}}{M_{i,n+1}}.
\end{split}
\label{eq:incDBIL}
\end{equation}

To calculate the index with the forgetting factor, \textit{DB}$_\lambda(n+1)$, we need to define $L_{\lambda i,n+1}$ to be used instead of $L_{i,n+1}$ and we control the decay of the denominator by clamping the decay at 1 with $\max$ function.
\begin{equation}
\begin{split}
L_{\lambda i,n+1} = \frac{C_{\lambda i,n+1}}{\max{\left\{1, M_{\lambda i,n+1}\right\}}}.
\end{split}
\label{eq:incDBILFF}
\end{equation}
\begin{figure*}[ht]
	\centering
		\subfigure[S1: Synthetic Locally Linear Processes]{
			\includegraphics[width=5.7cm]{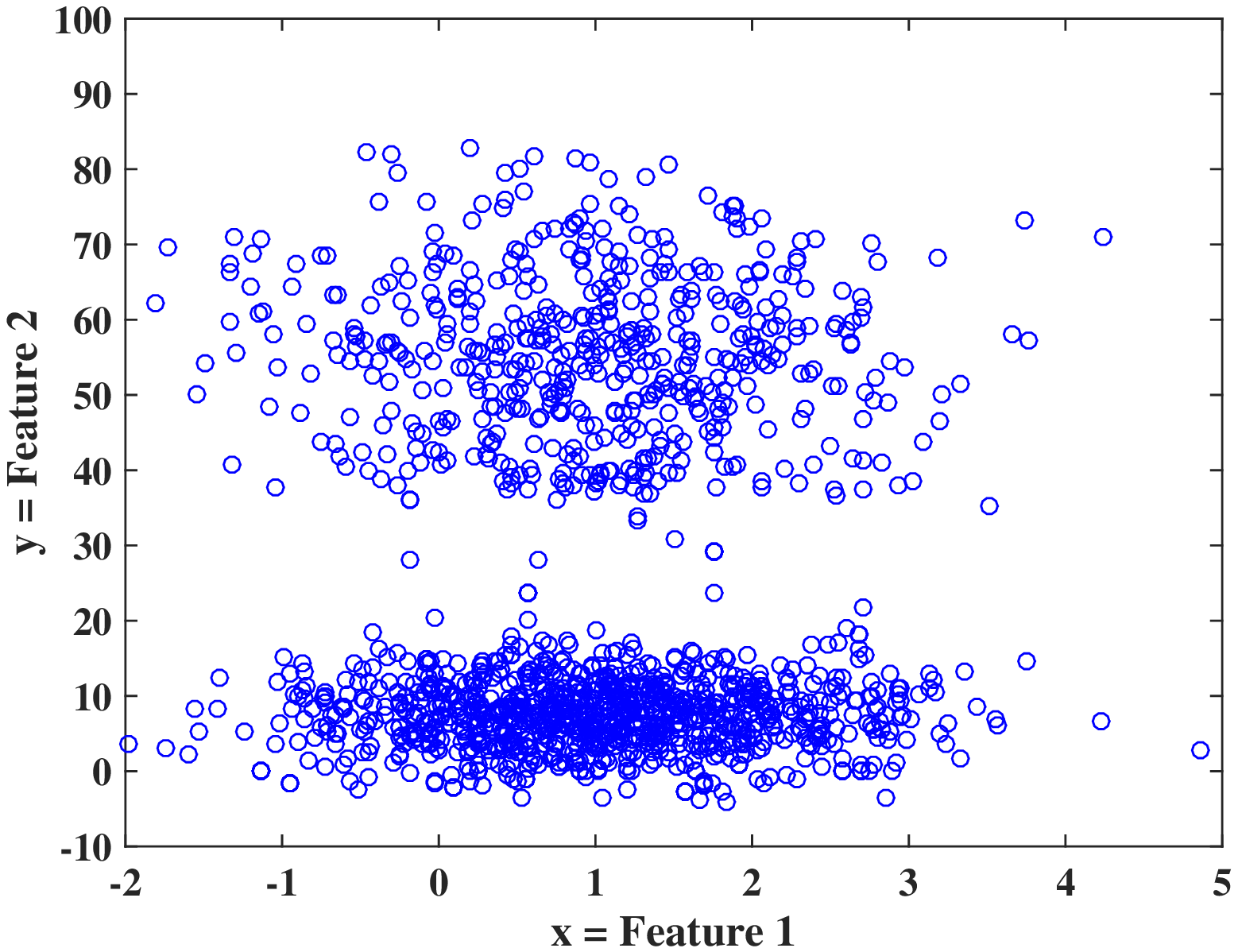}
			\label{fig:DSScattera}
		}
	  \subfigure[S2: Shifting Gaussian Distribution]{
			\includegraphics[width=5.7cm]{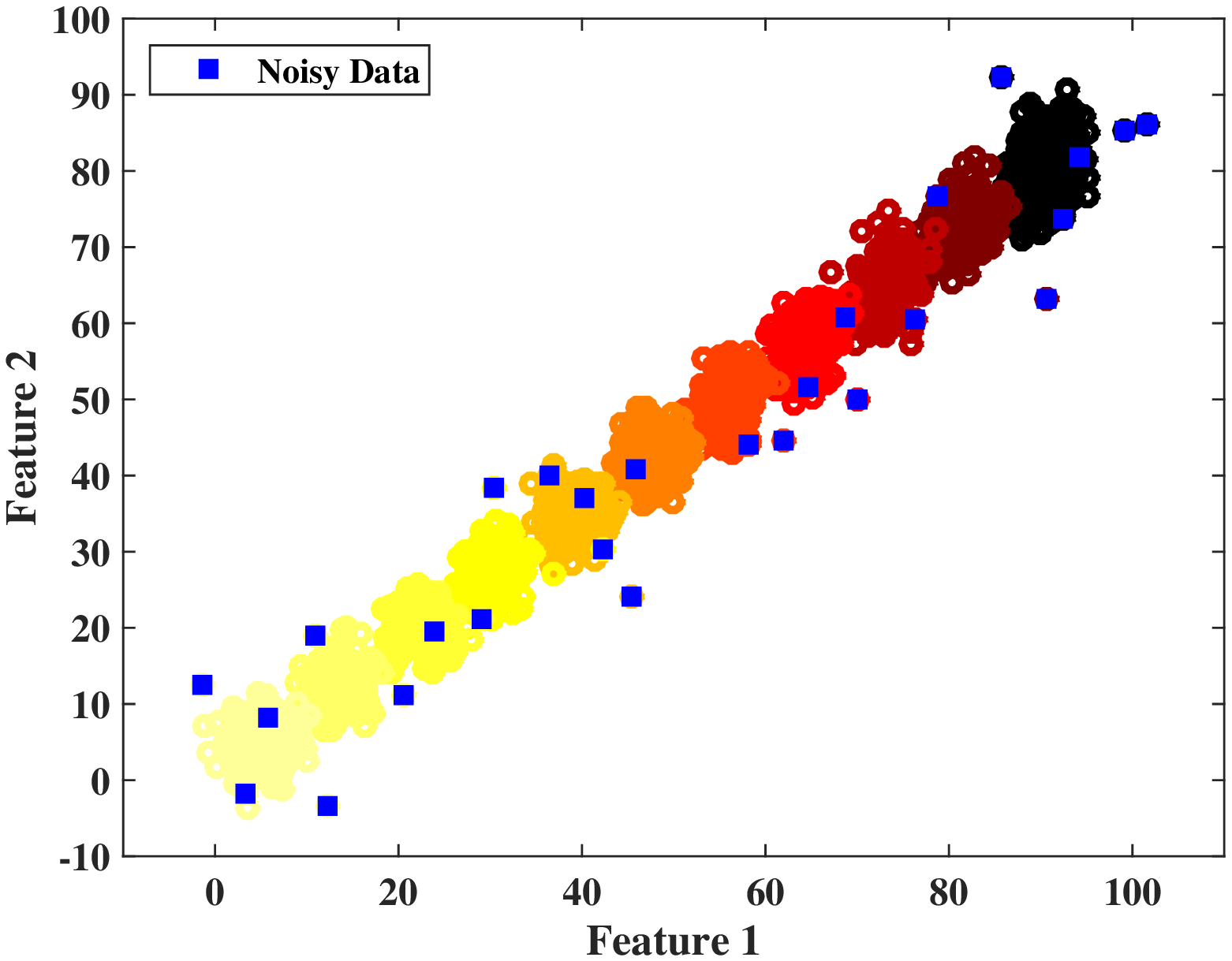}
			\label{fig:DSScatterb}
		}
	  \subfigure[S3: Circular Shifting Gaussian Distribution with Systematic Noise]{
			\includegraphics[width=5.7cm]{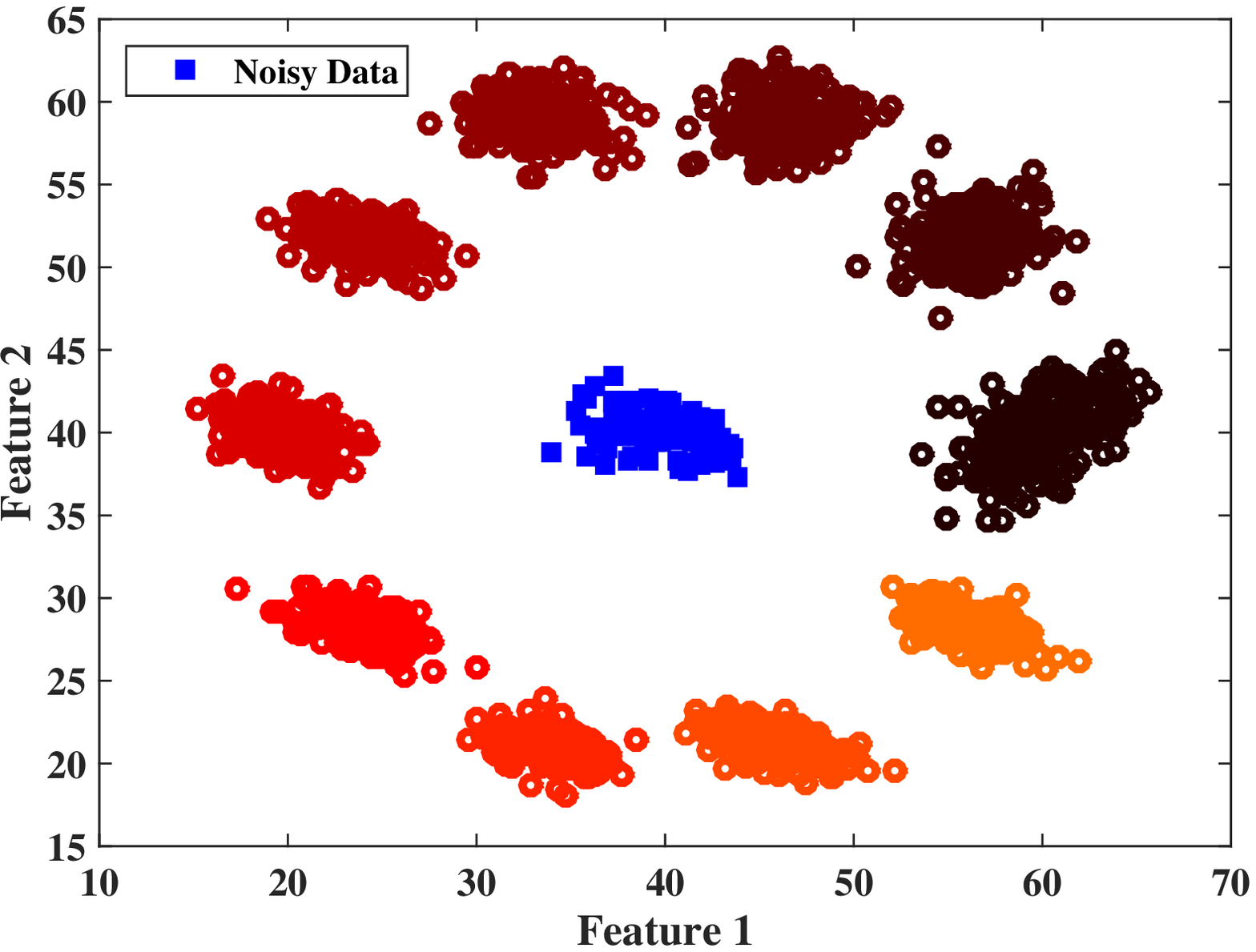}
			\label{fig:DSScatterc}
		}
		\caption{Scatter plots of three synthetic datasets.  S2 and S3 show the progression of time with colors, starting with black and becoming lighter with time.}
	\label{fig:DSScatterSynth}
\end{figure*}
\begin{table*}[ht]
	\centering
	\caption{Summary characteristics of the datasets used in the evaluations.}
	\begin{tabular}{m{1cm}|m{1.5cm}|m{1.5cm}|m{2.5cm}|m{2.5cm}|m{1cm}}
		Dataset&\multicolumn{1}{c|}{\# samples ($n$)}&\multicolumn{1}{c|}{\# Labels ($k$)}&\multicolumn{1}{c|}{\# dimensions ($p$)}&\multicolumn{1}{c|}{Noise Attributes}&\multicolumn{1}{c}{Labeling}\\ [0.5ex]
		\hline
		S1&\multicolumn{1}{c|}{1955}&\multicolumn{1}{c|}{2}&\multicolumn{1}{c|}{2}&\multicolumn{1}{c|}{No Noise - Drift between Clusters}&\multicolumn{1}{c}{Exact}\\ \hline
		S2&\multicolumn{1}{c|}{2727}&\multicolumn{1}{c|}{11}&\multicolumn{1}{c|}{2}&\multicolumn{1}{c|}{1\% - Uniform Random}&\multicolumn{1}{c}{Exact}\\ \hline
		S3&\multicolumn{1}{c|}{2000}&\multicolumn{1}{c|}{10}&\multicolumn{1}{c|}{2}&\multicolumn{1}{c|}{1\% - Systematic Random}&\multicolumn{1}{c}{Exact}\\ \hline \hline
		LG&\multicolumn{1}{c|}{2016}&\multicolumn{1}{c|}{3$^*$}&\multicolumn{1}{c|}{2}&\multicolumn{1}{c|}{Unknown}&\multicolumn{1}{c}{Our guess$^*$}\\ \hline
		GSA&\multicolumn{1}{c|}{9969}&\multicolumn{1}{c|}{3$^*$}&\multicolumn{1}{c|}{8}&\multicolumn{1}{c|}{Unknown}&\multicolumn{1}{c}{Our guess$^*$}\\ \hline
		\multicolumn{5}{l}{\small $^*$ These are unlabeled data sets, so we use physical arguments to justify the approximate} \\
		\multicolumn{5}{l}{\small number of clusters shown in column 2 for these two data sets.}\\
		\end{tabular}
	\label{tbl:Datasetsummary}
\end{table*}

As with the XB case, DB($n$) and DB($n+1$) are values of the incremental indices; iDB and iDB$_\lambda$𝛌 are the incremental models that produce them.


\section{Computational Protocols}
\label{sec:evaluation}
In this section, we first describe the synthetic and real-life datasets used in our evaluations and then we study the characteristics of iXB, $\text{XB}_{\lambda}(n)$, iDB and $\text{DB}_{\lambda}(n)$.
\subsection{Datasets and Parameters}
The synthetic dataset $S1$ consists of two-dimensional vectors $(x_q,y_q), q> 2$, which are generated using two modes, $M_1$ and $M_2$, with different dynamic functions and input signals ($x_{n}, q=3,4,\ldots$). Values of the independent variable ($x$) are random i.i.d. samples from a Gaussian distribution with $\mu=\sigma=1$. Values of the dependent variable ($y$) from $M_1$  and $M_2$ are then computed according to (\ref{eq:ARM1}) or (\ref{eq:ARM2}) respectively.

\begin{equation}
\begin{split}
& y_{n} = 1.018x_{n-1}+1.801y_{n-1}-0.8187y_{n-2} \\
& y_0=y_1=y_2=0
\end{split}
\label{eq:ARM1}
\end{equation}

\begin{equation}
\begin{split}
& y_{n} = x_{n-1}+0.5x_{n-2}+1.5y_{n-1}-0.7y_{n-2} \\
& y_0=y_1=y_2=0
\end{split}
\label{eq:ARM2}
\end{equation}

To build $S1$, we considered 4 mode changes between the two modes at uniform random intervals between 200 and 500 samples starting with $M_1$. Instead of a sudden shift between the modes, we gradually change the individual parameters of one mode to the other mode in 5 equal steps, during which we generate 10 samples in each intermediate mode. Fig.~\ref{fig:DSScattera} shows a scatter plot of the $S1$ dataset with the input (Feature 1 = $x$) and output (Feature 2 = $y$).

The second synthetic dataset $S2$ (shown in Fig.~\ref{fig:DSScatterb}), is generated by considering two modes, $M_1$ and $M_2$, with different two-dimensional normal distributions $N(\mu_1,\Sigma_1)$ and $N(\mu_2,\Sigma_2)$ and 9 intermediate modes. The parameter values of the modes $M_1$ and $M_2$ are: $\Sigma_1=\begin{pmatrix} 3.8418 & -2.6474 \\-2.6474 & 4.8478 \end{pmatrix}$, $\Sigma_2=\begin{pmatrix} 1.5239 & -0.5390 \\-0.5390 & 1.6467 \end{pmatrix}$, and  $\mu_1=(95,75)$ and $\mu_2=(5,5)$. 
$M_1$ is the initial mode, and $M_2$ is the final mode. $M_1$ is transformed as follows.
First, 500 samples $\left\{k=1\ldots500\right\}$ are drawn from $M_1$. Sampling continues as each individual value of the covariance matrix and the mean are changed in 10 equal steps from their values in $M_1$ to those in $M_2$. After the first step, 200 samples $\left\{n=501\ldots700\right\}$ are taken from the new normal distribution. After each new step 200 more samples are added to the dataset. The final step ends at mode $M_2$. The squares show 1\% of the samples from each normal distribution, which are perturbed by uniform noise from $[-10,\;10]$. A small level of noise is added to this dataset to investigate how the algorithms react to noise.

The third synthetic data set, S3, is generated by drawing samples from two-dimensional Gaussian distributions that rotate around a circle with 10 equal shifts. Before each shift, 200 samples are generated using the current Gaussian. In this data set, the noise (the blue cluster in the center of Fig.~\ref{fig:DSScatterc}) is generated by a Gaussian at the center of the circle. At each of 10 steps a random number of samples between 1 and 20 are removed from the outer distribution and then this number of samples are drawn from and added to the inner noise distribution. Table~\ref{tbl:Datasetsummary} summarizes the characteristics of the three synthetic datasets.

Column 2 of Table~\ref{tbl:Datasetsummary} specifies the number of physically labeled subsets in each data set. These subsets may, or may not, correspond to visually apparent or computationally acquired clusters. For example, if you imagine Figure 2(b) without the colors (which show the labels), most observers would assert that this data set has only 1 cluster.

We also use two real-life datasets. The LG dataset is from a collection of weather station nodes in the \textit{Le Genepi} (LG) region in Switzerland~\cite{WSN:GSB}. Two weeks of data at node 18 starting from October 10th 2007 are used in the evaluation. We use average surface temperature (T) and humidity (H) readings at node 18 over 10-minute intervals to form two dimensional input vectors $\left\{x_{n}=\left(T_n,H_n\right)\right\}$.
\begin{figure}[ht]
	\centering
			\subfigure[Scatterplot of LG]{
			\includegraphics[width=6cm]{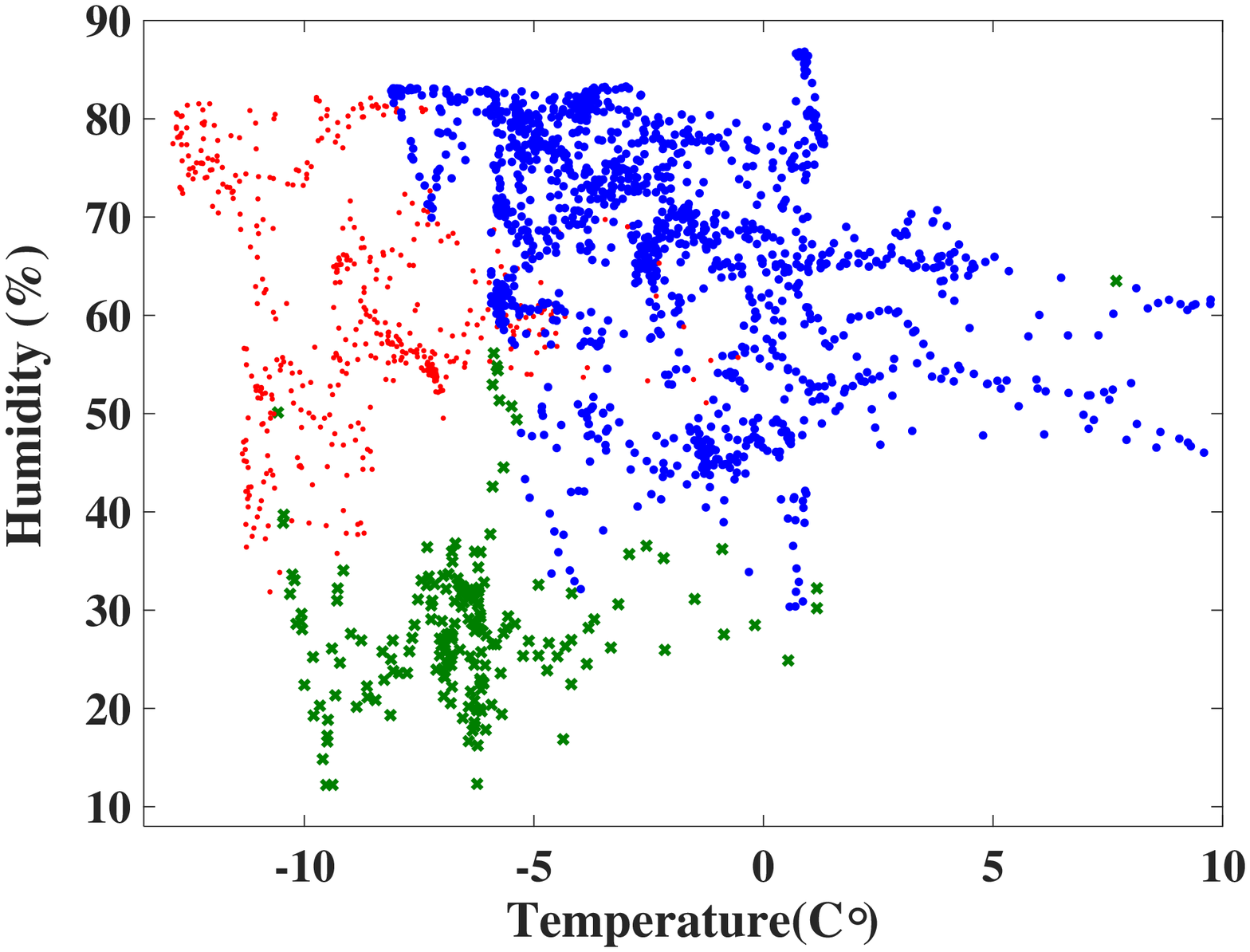}
			\label{fig:DSScatterReala}
			}\\
		\subfigure[Time-series view of the data]{
			\includegraphics[width=6cm]{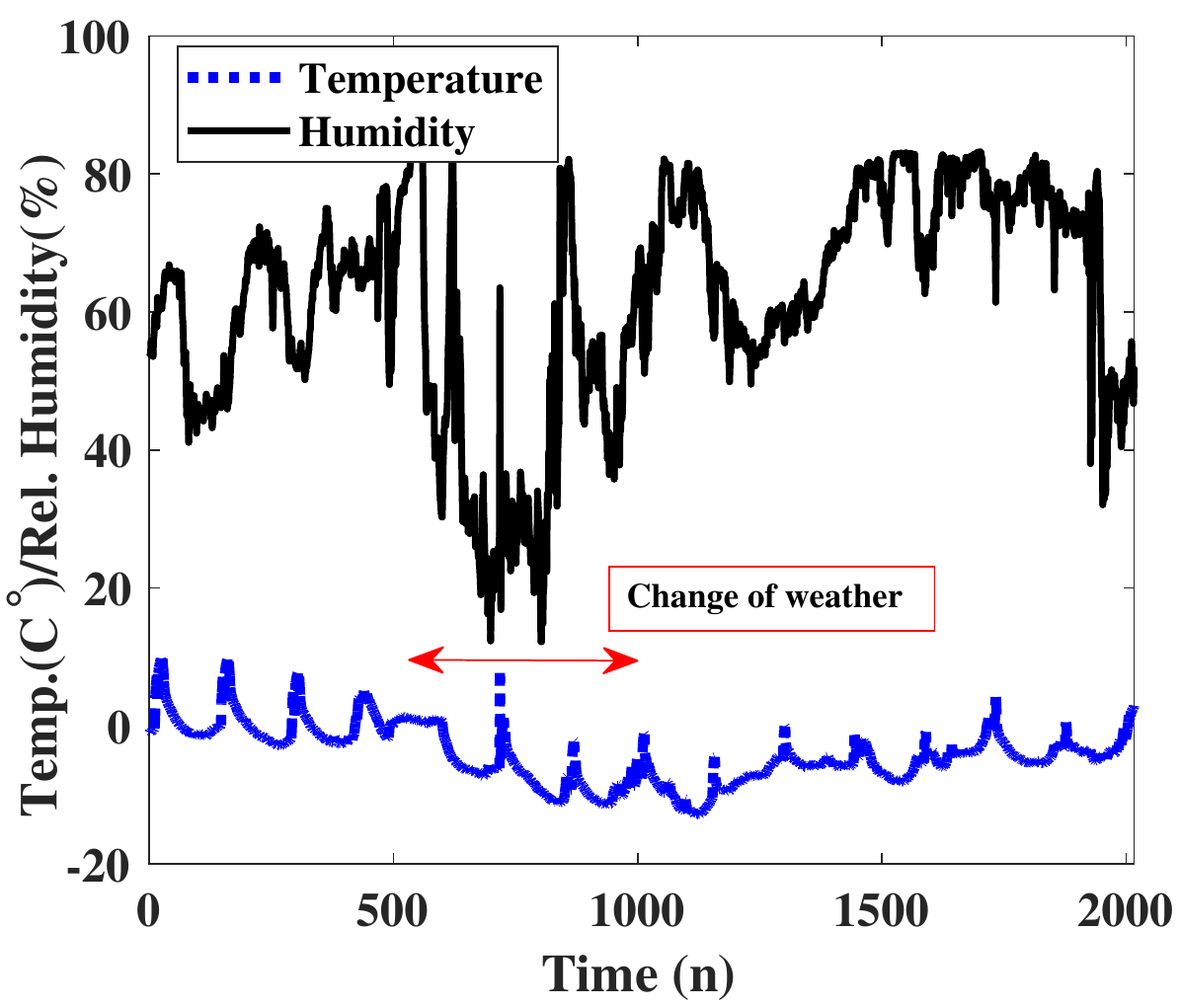}
			\label{fig:DSScatterRealb}
		}	
		\label{fig:DSScatterReal}	
		\caption{Scatterplot and time-series views of the LG data. The scatterplot shows three expected clusters.}
\end{figure}

Fig.~\ref{fig:DSScatterReala} shows a scatter plot of the LG data. The imagery information from the site shows that there is a snowy day during the two weeks of data collection and the data confirms that a cold and windy day precedes the snow. Fig.~\ref{fig:DSScatterRealb} shows this change of weather in time-series plots of temperature and relative humidity data. Therefore, we assume that there are three physical events, and that these may correspond to three clusters in the data: sunny days before and after the snow, cold front moving in, and the snowy day and label data accordingly, i.e., all of the points for days 1-6 and 10-14 are in cluster 1 (blue), day 7 in cluster 2 (green), and days 8 and 9 in cluster 3 (red). 

The second real-life data set is a gas sensor array under dynamic gas mixtures from the UCI repository that contains conductivity samples at 100Hz obtained from 16 chemical sensors (4 unique sensors). The conductivity of these types of sensors changes in the presence of different gas mixture concentrations. More information on the generated data set can be found in Fonollosa \textit{et al.}~\cite{Fonollosa2015}. We select two pairs of each unique sensor (8 sensors) and take the mean of their 100 samples per second over 5 minutes of the experiment as our evaluation data set (a stream of 300 8-dimensional input vectors), which we refer to as GSA. During this five minutes the sensors are exposed to two different concentrations of gases CO and Ethylene. This will lead to three distinct behaviours in the dataset: no gas being present, presence of CO and presence of Ethylene.

\subsection{Initialization}
Macqueen's sk-means algorithm has only one parameter to choose, $k$. We use the recommended parameter values for OEC from~\cite{OEC2016}. These parameters are forgetting factor $\lambda=0.9$, effective cluster boundary and outlier boundary threshold of 0.99 and 0.999 respectively, a stabilization period of $n_s=20$, and and an OEC forgetting factor 𝛌$\lambda_{\textit{OEC}}=0.9$.

The two clustering algorithms have slightly different initialization procedures. In sk-means the first $k$ points are the initial cluster prototypes and the index calculations start at the $k+1$-th point. In the OEC clustering algorithm, the first $p+1$ points are used to calculate a single cluster prototype and the cluster evaluation starts from point $p+2$ with only one cluster in the system.

Upon start of the evaluation at step $n$, the iCVIs are initialized with $C_{i,n}=0$, $M_{i,n}=n$ and $G_{i,n}= \overrightarrow{0}$ (zero vector in $\Re^p$). After initialization, the clustering algorithms and iCVIs process data one sample at a time.

\section{Numerical Experiments}
\label{sec:evaluation1}
We first study the effect of forgetting in the indices. In these experiments we specify $k$ to be the correct number of labels for sk-means for the synthetic datasets. We show that both indices with forgetting  reveal more information about the data streams than when forgetting is not in use.

\subsection{Forgetting or Not}
In the iCVI indices (similar to their batch counterparts) each new data point at time $n$ affects the overall value of the index with the weight $1/n$ (see (\ref{eq:iXB})). Therefore, over a long data stream, we expect $\textit{XB}(n)$ and $\textit{DB}(n)$ to become \textit{saturated} by data points and lose the sensitivity they need to reflect changes due to new data inputs.

\begin{figure}[ht]
	\centering{
			\includegraphics[width=6.3cm]{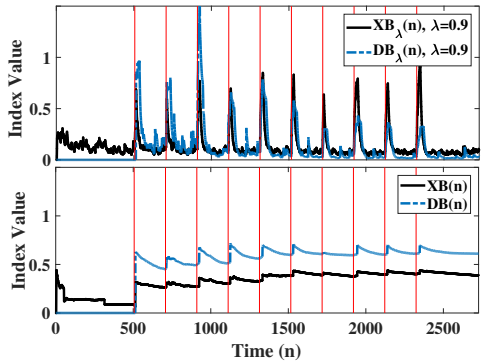}
		}
		\caption{Values of XB($n$), DB($n$), $\text{XB}_{\lambda}(n)$ and $\text{DB}_{\lambda}(n)$ with $\lambda=0.9$ in OEC clusters of the S2 dataset.}
	\label{fig:iXBNoForgettingOEC}
\end{figure}

Fig.~\ref{fig:iXBNoForgettingOEC} shows the values of the two indices with and without forgetting for online clustering in $S2$ with the OEC algorithm. The times when the distribution of samples changes are indicated with red vertical lines. You can see a sudden jump in the indices at these times. Even in  a fairly small data set such as $S2$, the jumps in the indices with no forgetting as $n$ increases (bottom of Fig.~\ref{fig:iXBNoForgettingOEC}) becomes very small. A data distribution change occurs around 1750, but neither DB($n$) nor XB($n$) show this change. This suggests that these two indices are not as suitable for monitoring evolving clusters in streaming data as their forgetting factor counterparts, which are shown in the top part of Fig.~\ref{fig:iXBNoForgettingOEC}. The jumps in the indices with forgetting factor are larger and much clearer than those in with no forgetting, and do not seem to depend on the number of samples processed by the index.


\begin{figure*}[ht]
	\centering
		\subfigure[S1: Synthetic Locally Linear Processes]{
			\includegraphics[width=5.7cm]{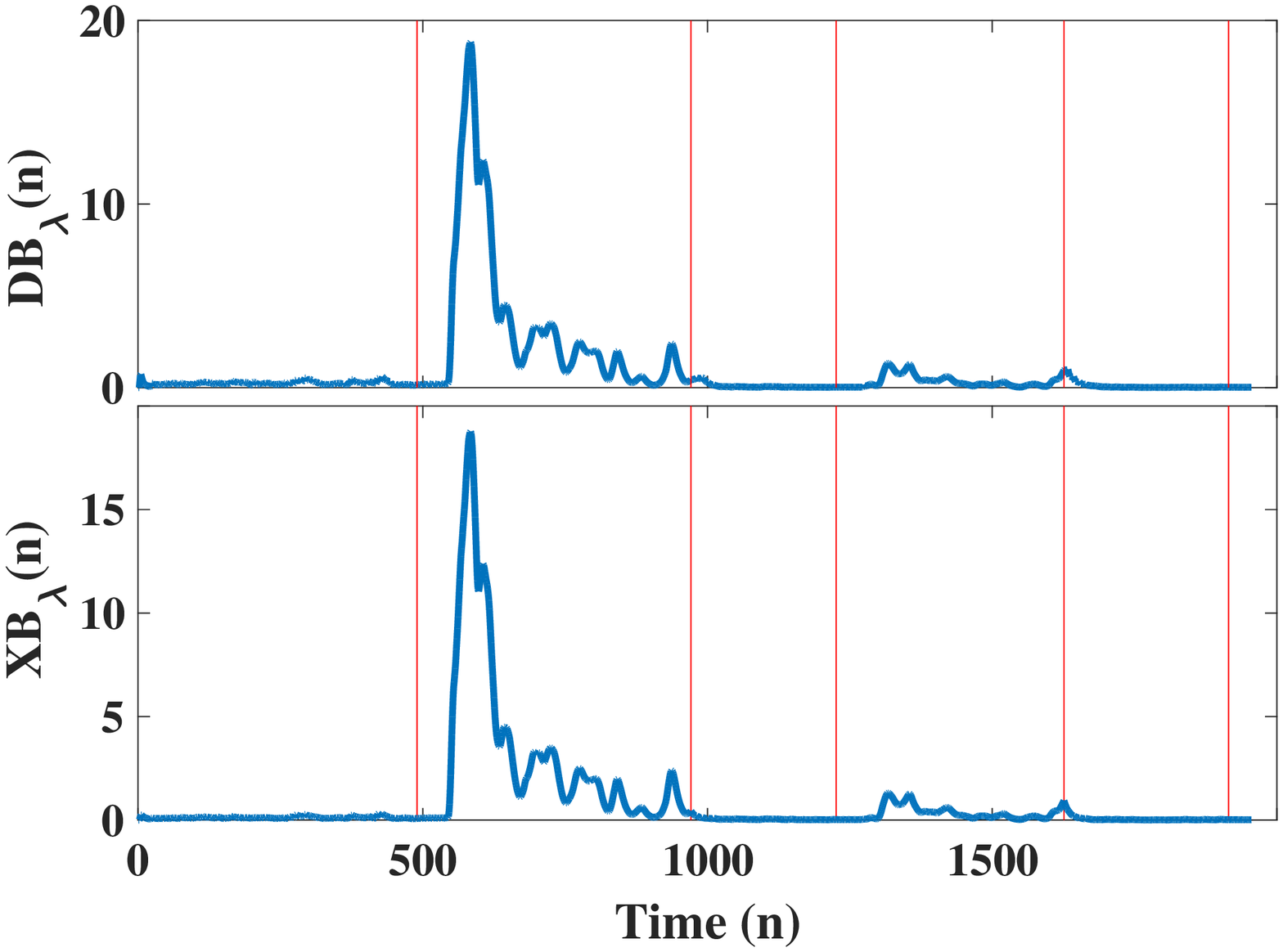}
			\label{fig:DSSynthiXBFFa}
		}
	  \subfigure[S2: Shifting Gaussian Distribution]{
			\includegraphics[width=5.7cm]{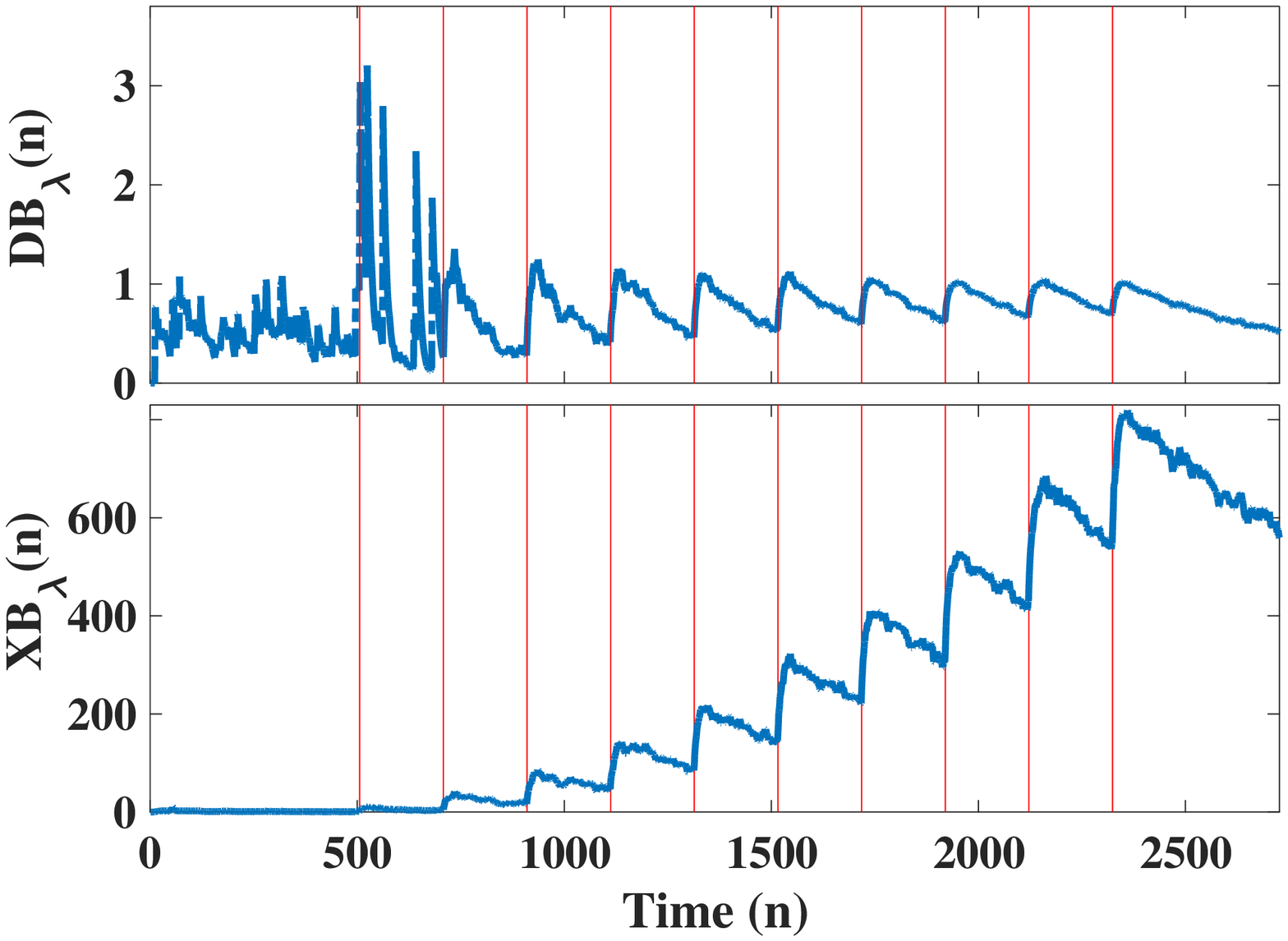}
			\label{fig:DSSynthiXBFFb}
		}
	  \subfigure[S3: Circular Shifting Gaussian Distribution with Systematic Noise]{
			\includegraphics[width=5.7cm]{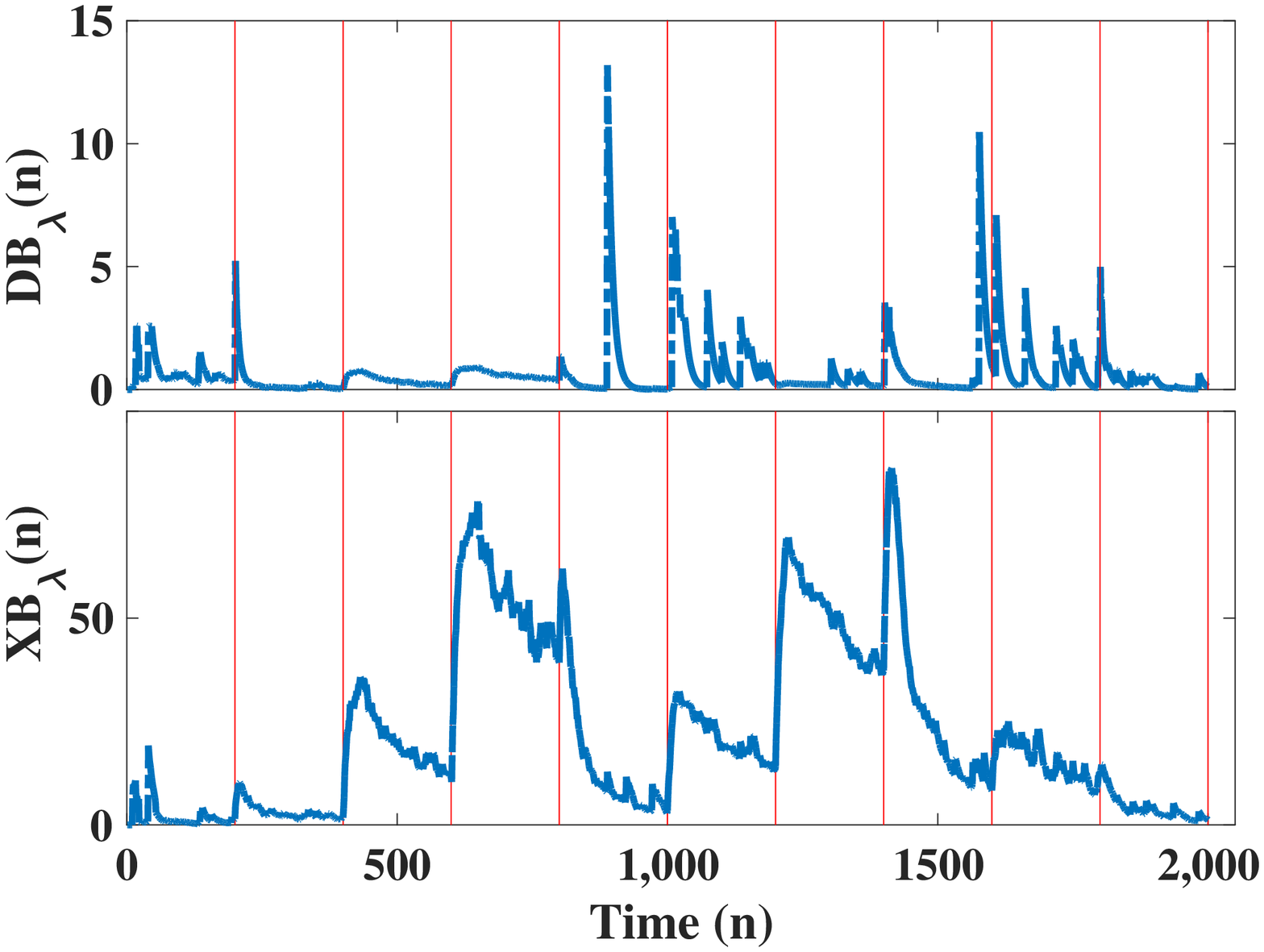}
			\label{fig:DSSynthiXBFFc}
		}
		\caption{Values of $\text{XB}_{\lambda}(n)$ and $\text{DB}_{\lambda}(n)$, $\lambda=0.9$, for sk-means. The vertical lines (red) are times when a change occurs in the dataset.}
	\label{fig:DSSynthiXBFF}
\end{figure*}

\subsection{Interpretation of the iCVI model output over time}
To study the usefulness of XB$_{\lambda}(n)$ and DB$_{\lambda}(n)$ in interpreting the results of online clustering algorithms, we first plot their values over time in clusters found by sk-means in the three synthetic datasets. In most cases because of the larger number of clusters and lower accuracy of sk-means in finding the expected clusters, the scale of the iCVI values for the sk-means approach is much higher than the values corresponding to OEC clusters. Therefore, we show the changes of $\text{XB}_{\lambda}(n)$ and $\text{DB}_{\lambda}(n)$ over time for only sk-means in Fig.~\ref{fig:DSSynthiXBFF}. We will discuss the noteworthy trends in OEC using separate figures.

In Fig.~\ref{fig:DSSynthiXBFF}, The indices $\text{XB}_{\lambda}(n)$ and $\text{DB}_{\lambda}(n)$ have almost identical performance for data set $S1$, as can be seen by comparing graphs of their values in Fig.~\ref{fig:DSSynthiXBFFa}. But their performance is quite different for data sets $S2$ and $S3$, as can be seen in the top and bottom views in Figs.~\ref{fig:DSSynthiXBFFb} and \ref{fig:DSSynthiXBFFc}, respectively. We highlight two main traits in values of these indices. These traits relate to the appearance of new clusters in the stream (spikes in the index) and performance of the clustering algorithm in modeling the new cluster (the reduction of the index after a spike).

\subsubsection{Appearance of new clusters}
CVIs assess cohesion and separation so sudden changes in the value of a CVI indicate changes in the cohesion and separation of the clusters produced by the clustering algorithm. Thus, a sudden change in an online validity index usually indicates the appearance of a new cluster in the data stream. This superficially relates to change-point detection in data streams but the distinction needs to be made that iCVIs monitor the partitions produced by the clustering algorithm and do not search for change-points in the streams. However, change-points that result in appearance of new clusters in the data streams change the cohesion of the clusters and cause spikes in the iCVI values.

The vertical lines in Fig.~\ref{fig:DSSynthiXBFF} mark the times where new clusters appear in the synthetic datasets. The sk-means algorithm performs reasonably well on the $S1$ dataset, completely fails for the $S2$ dataset (Fig.~\ref{fig:DSSynthSKMa}), and partially identifies the evolving clusters in $S3$ (Fig.~\ref{fig:DSSynthSKMb}). As we shall see, the OEC algorithm performs reasonably at detecting times when new distributions are created in all three datasets.

There is a jump in the values of both indices in Fig.~\ref{fig:DSSynthiXBFF} shortly after each new cluster is introduced in the data stream. As the algorithms learn the prototype that represents the new distribution in the data, the value of the index drops (bear in mind that these indices are all min-optimal). This behaviour is clearly shown in $\text{XB}_{\lambda}(n)$ plots but the $\text{DB}_{\lambda}(n)$ plot in the top view of Fig.~\ref{fig:DSSynthiXBFFc} does not reflect this behaviour. We attribute this to partial identification of clusters in $S3$ and the systematic noise in this dataset which affects $\text{DB}_{\lambda}(n)$ more than $\text{XB}_{\lambda}(n)$. 



\subsubsection{Distress signals in the clustering algorithm}
\begin{figure}[ht]
	\centering
	  \subfigure[S2: Shifting Gaussian Distribution]{
			\includegraphics[width=5.7cm]{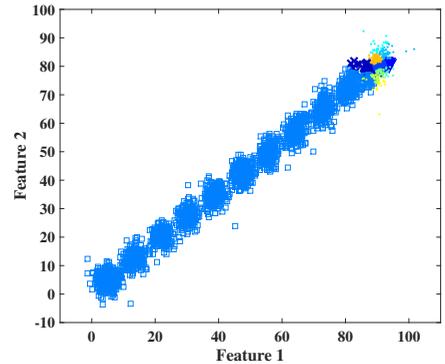}
			\label{fig:DSSynthSKMa}
		}
	  \subfigure[S3: Circular Shifting Gaussian Distribution with Systematic Noise]{
			\includegraphics[width=5.7cm]{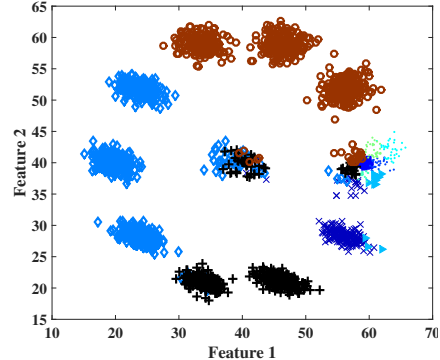}
			\label{fig:DSSynthSKMb}
		}
		\caption{The terminal clusters produced by sk-means in $S2$ and $S3$}
	\label{fig:DSSynthSKM}
\end{figure}

Another valuable asset of the graphs of incremental validity indices with forgetting is that a streaming plot of their values can exhibit signs of failure. The sk-means algorithm performs well in the $S1$ data set and we can see a sharp peak (sudden increase and decrease) in Fig.~\ref{fig:DSSynthiXBFFa} after the first time the second cluster appears in the data. The $\text{XB}_{\lambda}$ and $\text{DB}_{\lambda}$ indices are almost identical in this data set. However, sk-means fails to identify the expected clusters in both $S2$ and $S3$ as shown by the end-state partitions in Fig.~\ref{fig:DSSynthSKM} even when the correct number of clusters are supplied to the algorithm. Let's see how the $\text{XB}_{\lambda}(n)$ plots reflect this problem in the sk-means clustering algorithm.

Fig.~\ref{fig:DSSynthiXBFFb} shows $\text{XB}_{\lambda}(n)$ over time and the final clusters in $S2$ shown in Fig.~\ref{fig:DSSynthSKMa} by sk-means. The $\text{XB}_{\lambda}(n)$ plot shows that its values have an increasing trend. In this dataset, we have 11 clusters that appear one by one in the stream. We know that the appearance of new clusters results in spikes in the $\text{XB}_{\lambda}(n)$ values. After the appearance of a new cluster in the data, we expect the clustering algorithm to create a prototype for the cluster and update the partition to account for the newly observed cluster. Subsequently, this should bring any min-optimal \textit{down} to similar (or lower) values that were observed before the new cluster appeared. The total failure of sk-means in $S2$ is evident by its inability to restore the $\text{XB}_{\lambda}(n)$ values after a new cluster enters the data stream. Indeed, increasing values of these min-optimal indices signals that things are going awry! The $\text{DB}_{\lambda}(n)$ values show smaller peaks and decreases but still shows that sk-means cannot find good clusters to sharply reduce the index. The graph in the top view of Fig.~\ref{fig:iXBNoForgettingOEC} shows the values of both indices for the OEC algorithm applied to $S2$, which identifies all the clusters correctly. The sharp peaks in the values verifies the fact that the clustering algorithm finds all the expected clusters.


Figs.~\ref{fig:DSSynthiXBFFc} and \ref{fig:DSSynthSKMb} tell a slightly different story for $S3$. The sk-means clusters in this dataset provide some separation between the expected clusters. The separation is achieved by creating two partitions with 3 clusters each, a partition with two clusters and the final cluster is identified correctly. At two points during the experiment, there are significant reductions in the $\text{XB}_{\lambda}(n)$ values (around 800 and 1500 samples) which correspond to the times after the two subsets of 3 clusters are created. The decreasing trend of $\text{XB}_{\lambda}(n)$ towards the end is due to the fact that sk-means performs slightly better in identifying the last 3 clusters. In particular, correct identification of the last cluster reduces the index value quite sharply. The $\text{DB}_{\lambda}(n)$ values in this dataset are very different to the $\text{XB}_{\lambda}(n)$ values. The index $\text{DB}_{\lambda}(n)$ generates large peaks around the noisy data and shows higher sensitivity to systematic noise in the data. 

Sharp peaks corresponding to sudden drops in $\text{XB}_{\lambda}(n)$ after the spikes (appearance of a new cluster) relates to the clustering algorithm appropriately creating a prototype for the cluster, while a gradual decrease after a spike is a sign of the failure of the algorithm to identify the new cluster. The $\text{DB}_{\lambda}(n)$ index is more sensitive to large amplitude local noise and generates smaller peaks for the new clusters. This hinders the interpretation of the clustering results in terms of appearance of new clusters and learning problems in the algorithm.

\begin{figure}[ht]
	\centering{
			\includegraphics[width=6.3cm]{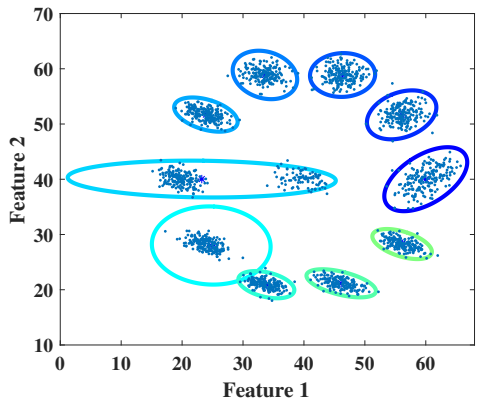}
		}
		\caption{Terminal OEC clusters on $S3$ dataset.}
	\label{fig:OECClusters}
\end{figure}

\begin{figure}[ht]
	\centering
			\includegraphics[width=5.7cm]{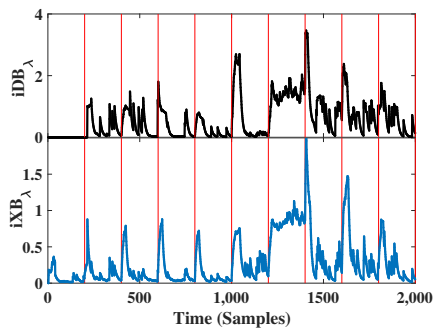}
		\caption{$\text{XB}_{\lambda}(n)$ and $\text{DB}_{\lambda}(n)$ values for OEC clusters in $S3$}
	\label{fig:DSSynthOECForced}
\end{figure}

So far we only looked at problems in the sk-means algorithm because OEC finds all the expected clusters in these three synthetic datasets. However, In $S3$, as shown in Fig.~\ref{fig:OECClusters}, one of the cluster prototypes has been expanded and covers the systematic noise in the centre of the plot (The expanded prototype is a single prototype that represents the two clusters of data captured by the horizontally elongated ellipse in Fig.~\ref{fig:OECClusters}). Fig.~\ref{fig:DSSynthOECForced} shows the values of the two indices for OEC clustering in this dataset. Both indices have similar trends with $\text{DB}_{\lambda}(n)$ producing larger peaks for noisy data. Neither graph in Fig.~\ref{fig:DSSynthOECForced} experiences a drop after creating this prototype. This indicates that OEC does not produce a good model of the data which matches the visual assessment of the clusters.

The overall conclusion from our synthetic data experiments is that $\text{XB}_{\lambda}(n)$ is much more effective than $\text{DB}_{\lambda}(n)$, so the remaining discussion will involve only $\text{XB}_{\lambda}(n)$.


\subsection{Experiments with real data}
\begin{figure}[ht]
	\centering
	  \subfigure[LG: Sensorscope project in Le Genepi]{
			\includegraphics[width=5.7cm]{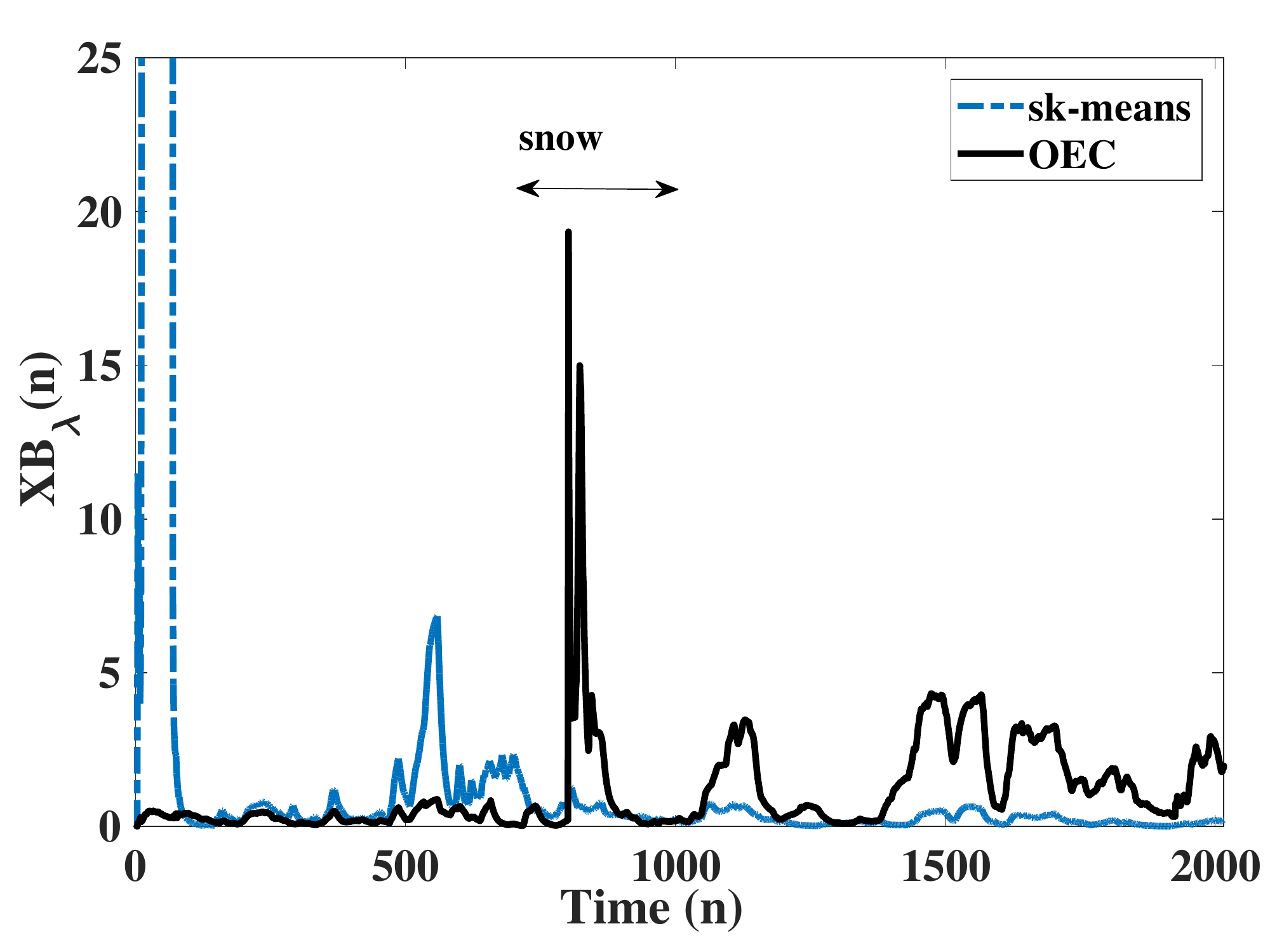}
			\label{fig:iXBReala}
		}
	  \subfigure[GSA: Gas Sensor Array. Vertical lines indicate the times where concentration of gases change]{
			\includegraphics[width=5.7cm]{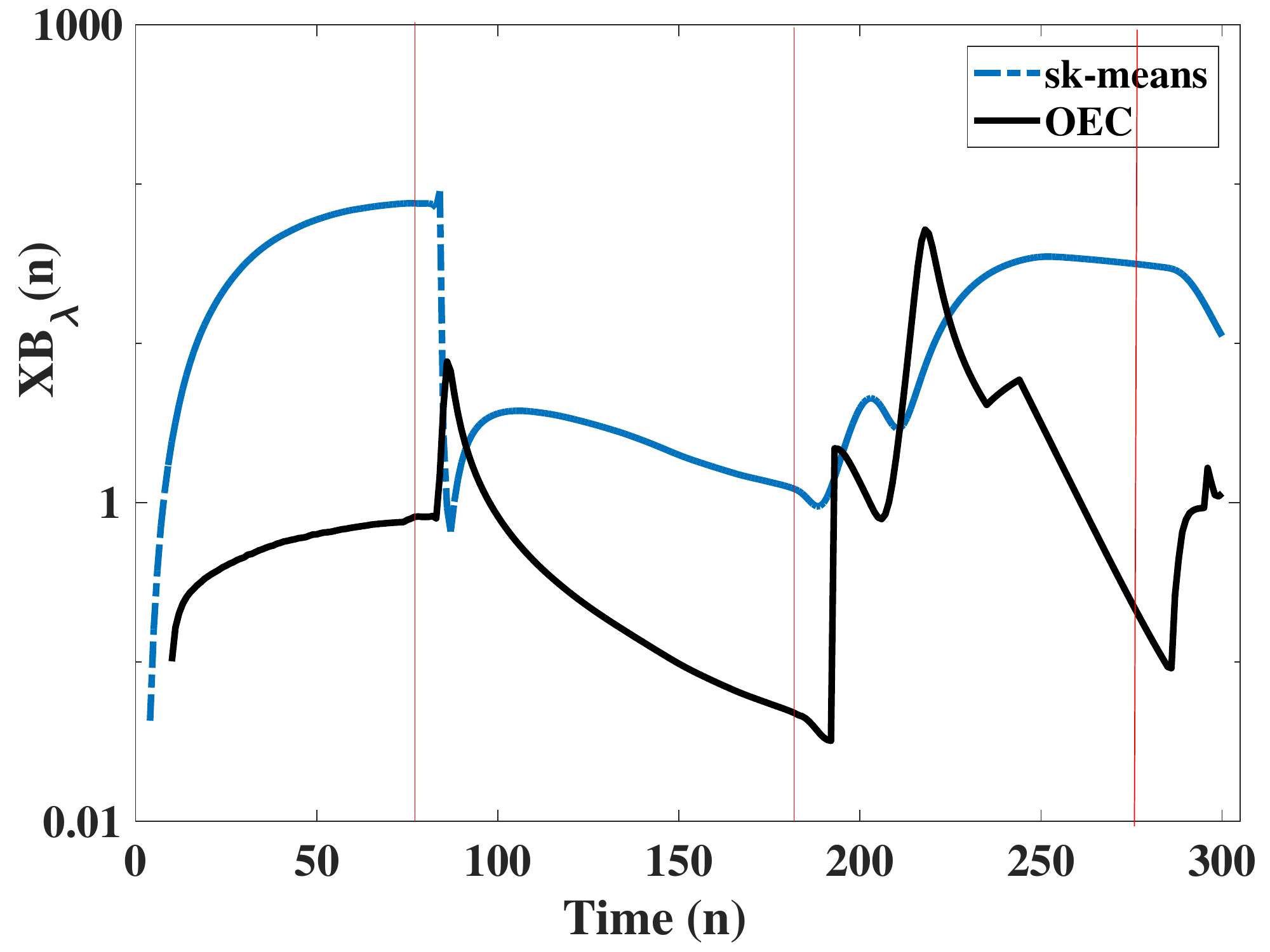}
			\label{fig:iXBRealb}
		}
		\caption{The $\text{XB}_{\lambda}(n)$ index over the real-life datasets. The y-axis of GSA plot is shown in logarithmic scale.}
	\label{fig:iXBReal}
\end{figure}

Since we have only assumed information about the ground truth in the real datasets it is harder to interpret the $\text{XB}_{\lambda}(n)$ values. We will use the insights obtained from our analysis of the synthetic datasets to understand the behaviour of $\text{XB}_{\lambda}(n)$ when clustering in the real-life datasets. Fig.~\ref{fig:iXBReala} shows values of $\text{XB}_{\lambda}(n)$ calculated for both OEC and sk-means (with $k=3$ clusters) in the LG dataset. The major event in this dataset is a snowfall event (the approximate location of the event between about $n=700$ and $n=900$ is marked in the figure). OEC identifies three clusters in this dataset~\cite{OEC2016} which correspond to normal days, high wind before the snow and the snow. The two very close sharp spikes followed by drops in both indices values corresponds the spike for the wind before the snow and the snow fall as identified by OEC. The first spike corresponds to the time when the high wind had become the dominant feature in the area. Some level of wind can be observed from step 600.

The index values for sk-means clusters do not correspond to the major events in the dataset. There is only one major peak around the very start of the high winds at about step 600. Since sk-means does not consider the correlation in time between the data points, it fails to account for the temporal nature of the clusters in the LG data. As shown in Fig.~\ref{fig:DSScatterReal}, we must to look at the evolution of data in time to see the major events and the scatter plot of the data does not show a clear cluster tendency. The large overlap between clusters of data corresponding to major events leads to the fact that aligning clusters with the events does not guarantee the maximal separation. This is reflected in the index values for sk-means in Fig.~\ref{fig:iXBReala} that shows no clear increasing trend or particularly slow reduction in the index values. However, the index values for sk-means clusters are mainly lower than those for OEC clusters showing that sk-means clustering creates relatively better clusters in terms of cohesion and separation.


The GSA dataset is collected in a more controlled environment than LG and hence, has a more recognizable cluster tendency. In this dataset two different gas concentrations are introduced during the experiment. When the gas is introduced (times indicated by the red vertical lines) there is a delay until the sensors react to the presence of the gas, which is seen as a delay of the values in Fig.~\ref{fig:iXBRealb}.  The peaks in both indices occur a few seconds after the introduction of the gas. For the first event at about $n=75$, a gas is introduced when no other gas is present, while the second event at about $n=180$ corresponds to removing one gas and introducing another gas. In the second case, the indices show two peaks close to each other for both clustering algorithms. The last event at about $n=270$ corresponds to emptying the chamber and since the empty chamber had been seen by the clustering algorithms, this results in a much smaller peak in the graphs. OEC has smaller $\text{XB}_{\lambda}(n)$ values than sk-means, which indicate that in this dataset, OEC results in a better separation between the expected clusters than sk-means.

\section{Conclusions}
\label{sec:conclusion}
To the best of our knowledge, the concept of incremental validity indices (iCVIs) in data stream clustering has not been studied before. Our iCVIs offer an unsupervised validation mechanism for online clustering algorithms. In this article, we introduced the novel concept of online iCVIs and derived forgetting and non-forgetting versions of two well-known internal validity indices (the Xie-Beni indices (XB($n$) and $\text{XB}_{\lambda}(n)$) and the Davies-Bouldin indices (DB($n$) and $\text{DB}_{\lambda}(n)$)). Our experiments used two different styles of algorithms for streaming clustering: sk-means, which does not account for the historical context of streaming data; and OEC, which retains a history of time dependency in clusters through the retention of its cluster statistics (means and covariances). We discussed how these indices can be used to interpret the performance of online clustering algorithms with respect to the appearance of new clusters and how the clustering algorithm reacts to evolving clusters. An increasing trend in any min-optimal index seems to point to a more questionable belief about cluster quality in the clustering of
the data. Thus, our iCVI models afford a means for sending distress signals about evolving clusters to real time monitors. 


Our experiments indicate that the most reliable of the four incremental indices \{XB($n$),$\text{XB}_{\lambda}(n)$, DB($n$),$\text{DB}_{\lambda}(n)$\} is $\text{XB}_{\lambda}(n)$. And of the two clustering algorithms used, OEC seems to perform much better than sk-means. But definitive conclusions require many more tests. There are many, many internal CVIs. Our next focus will be on deriving other incremental validity indices and conducting comparative studies among different types of indices.

\section{Acknowledgment}
This research was supported under Australian Research Council's \textit{Discovery Projects} funding scheme (project number DE150100104).

\end{document}